# Salient Positions based Attention Network for Image Classification*

Sheng Fang†  
College of Computer Science and Engineering  
Shandong University of Science and Technology  
China  
fangs99@126.com

Kaiyu Li  
College of Computer Science and Engineering  
Shandong University of Science and Technology  
China  
likyoo@sdust.edu.cn

Zhe Li  
College of Computer Science and Engineering  
Shandong University of Science and Technology  
China  
lizhe@sdust.edu.cn

## ABSTRACT

The self-attention mechanism has attracted wide publicity for its most important advantage of modeling long dependency, and its variations in computer vision tasks, the non-local block tries to model the global dependency of the input feature maps. Gathering global contextual information will inevitably need a tremendous amount of memory and computing resources, which has been extensively studied in the past several years. However, there is a further problem with the self-attention scheme: is all information gathered from the global scope helpful for the contextual modelling? To our knowledge, few studies have focused on the problem. Aimed at both questions this paper proposes the salient positions-based attention scheme SPANet, which is inspired by some interesting observations on the attention maps and affinity matrices generated in self-attention scheme. We believe these observations are beneficial for better understanding of the self-attention. SPANet uses the salient positions selection algorithm to select only a limited amount of salient points to attend in the attention map computing. This approach will not only spare a lot of memory and computing resources, but also try to distill the positive information from the transformation of the input feature maps. In the implementation, considering the feature maps with channel high dimensions, which are completely different from the general visual image, we take the squared power of the feature maps along the channel dimension as the saliency metric of the positions. In general, different from the non-local block method, SPANet models the contextual information using only the selected positions instead of all, along the channel dimension instead of space dimension. The various experiments in CIFAR and Tiny-ImageNet datasets show the proposed method can outperform the non-local block even with a far smaller affinity matrix, especially when they are used in the low network levels. Our source code is available at https://github.com/likyoo/SPANet.

## CCS CONCEPTS

• Computing methodologies~Machine learning~Machine learning approaches~Neural networks   • Computing methodologies~Artifi-cial intelligence~Computer vision~Computer vision problems ~Object recognition

## 1 Introduction

In the past several years, attention mechanism has played an important role in deep learning for capturing long-range dependencies among the input data. Since attention is firstly integrated into neural network for natural language processing [1], it has been widely proved to be effective in various deep learning task, such as machine translation [2], language modeling [3], image classification [4,5], semantic segmentation [6,7], visual question answer [8], image captions [9], etc. More and more various attention modifications have been reported and enrich attention mechanisms. These attention mechanisms applied in computer vision tasks could be roughly divided into three categories.

The first class of attention mechanism is spatial attention, which originates from the self-attention of modeling language long dependency and acquires SOTA performance [3]. The self-attention is used firstly in scene segmentation as the name of non-local block [5] and also achieves the SOTA scores in various compute vison tasks. This scheme often uses three transforms to generate the query, key and value tensors based on the input feature maps, and its output involves the traverse operations in the three tenors. So the self-attention scheme is often criticized for its tremendous demands for computing and memory resources. One approach to solve this problem involves using less data to compute the query-key scores for sparing a great amount of computing [10,11,12,13,14,15], and certainly most of them report a slight performance decrease. However, there is a further problem with self-attention: does all the dependency generated within the global scope be beneficial? In the top two rows of Figure 1, it could be noticed that some texture in background, e.g. the texts in top middle of the images, also lead to the high responses in attention maps. Evidently it is negative for any computer vision tasks. As far as we know, in modeling the long dependency by self-attention scheme, few previous researches have investigated the problem of distinguishing between the positive and negative information.

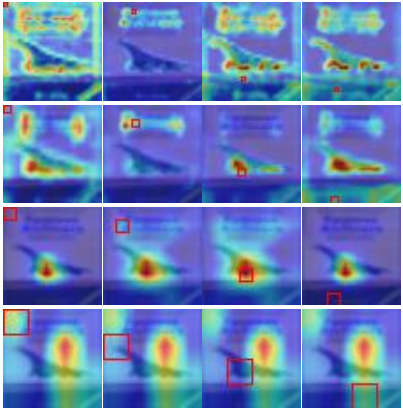

Figure 1: The attention maps for different query positions (denote by the red box) generated by NL-block [5] in ResNet-18 [32]. From top to bottom are the maps belonging to stage 1, 2, 3, and 4, with size of 32×32, 16×16, 8×8, and 4×4 respectively, and for visualizing all resized to 56×56. The image is from CIFAR datasets [33].

The second category integrating attention into neural networks is channel attention. One typical work of this class is SENet [4], which uses an average pooling of input feature maps following space dimension to recalibrate the channel. The design of SENet [4] is simple, concise, but at the cost of a great increase of network parameters for two fully connected layers. Aimed at this question, ECA [16] proposes to use 1d convolution instead of 2d convolution to carry out the excitation, and the kernel size of conv1d could be adaptively set. This will decrease a much amount of the parameters and the accuracy of image classification is even improved in comparison with SENet for keeping the channel number unreduced. In addition, the works [17,18,19,20,21] have studied using various information in the process of recalibrating the channel, and more or less they could improve the channel attention performance.

The work [12] calls the self-attention [1,3,5] as dot-product attention while the channel attention [4] as scaling attention, and regards them as two completely separate sets of techniques with very different goals. The former class stresses on modeling the global dependency, and the latter one focuses on recalibrating the channel attention in order to emphasize the important features. We argue that the two goals have intersections and have the chance to fuse together. In fact, [22] has proposed to combine the channel attention and spatial attention based on the observation that in different query positions the self-attention attention maps almost are the same. However, the observation keeps true only when self- attention is used in the high levels of neural networks. As the bottom two rows of Figure 1 show the attention maps generated in the stage 3 and 4 are very similar to each other, but the top two rows of Figure 1 demonstrate that the attention maps produced in the low level networks may be greatly different from each other. Additionally, [23] proposes to utilize the query, key and value tensor [3] not only in space dimension but also in channel dimension for segmentation, of course this will lead to more computing complexity.

Inspired by above works, especially [5, 22], we propose a Salient Positions based Attention Network (SPANet), which tries to select only the salient positions to attend the computing of query-key scores and pool the information through the channel dimension instead of space dimension. The motivations of the proposed scheme come from two observations. The first observation is that although in many queries positions the attention maps are similar, there are many different attention maps w.r.t the queries positions, and whether they are similar is highly correlated with the network stages or other factors, as shown in Figure 1. The second observation comes from the fact the salient area in attention maps often come from the query positions on the textures or the edges, especially in the low level network layer with exact spatial information, as shown in Figure 2 and Figure 3. Both observations lead us to the proposed SPANet, which explores to produce the attention maps in salient query positions for gathering contextual information. This approach decreases the computing complexity and memory for using less data, and at the same time tries to distill the positive contextual information by selecting the salient positions for the performance improvement. As far as we know there are few works studied both questions.

The principle contributions of this paper are summarized as below:

(1) We present the SPANet not only to decrease the need of computing and memory resources but also to distill the positive contextual information for producing the attention maps

(2) A Salient Positions Selection (SPS) algorithm is designed especially for the feature maps produced in neural networks, for they are handled hardly for their high channel dimensions.

(3) The SPANet firstly selects the salient positions according to the self-attention style, and then different from the self-attention scheme pools and models the information along the channel dimension.

(4) More than the complex objects such as CIFAR, the simple ones with clean background like MNIST are explored, and some observations are reported, which will be a beneficial complement for the previous works [22,35,36].

The rest of this paper is organized as follows. We first review the related works in Section 2 and present the proposed methods in Section 3. More observations and experimental results are presented and analyzed in Section 4. In the final section come our conclusions and future works.

## 2 Related works

### 2.1 Salient area/object detection

Salient object detection (SOD) tries to highlight visually salient object regions in images. The 'visually salient' means it is attractive to human [24]. So the reported SOD techniques often target to the visual image with three or one channels. In the deep learning era, the SOD methods often use neural architecture [24], and in general suffer from the need of great train data with ground truth. So it is not easy to plug these works into our network. Some earlier SOD works don't need train data, but they mainly depend on the low level features, such as the contrast in chrominance, motion, and intensity [25]. This means when the statistical property of the data changes these works will be no longer in force. The work [26] detects the salient areas according to the L2 norm of the difference between the mean image average vector and the corresponding image pixel vector value after the Gaussian blurred operation. Aimed at the detection of the salient object in gray or colorful image, [27] proposes two algorithms based on the intensity or color contrast by mapping the larger intensity/color space to a smaller one. However, in this paper we will deal with the feature maps produced in the neural network. At one hand they have larger channel dimensions than the visual images, e.g. in ResNet18 the feature maps often have 64 channels only after the first stage. At the other hand, the coefficients of the feature maps evidently show different statistical characters from the visual images. So apparently the previous SOD methods suitable for visual images are not appropriate for the salient positions detection for the neural networks feature maps.

## 2.2 Contextual information aggregation

For channel attention mechanism, it is a common practice to aggregate as much contextual information as possible to augment the feature representation. Many studies have been done by exploiting more information from various sources. The work [28] proposes to add the spatial pyramid by the average pooling with three different sizes, in contrast [17] uses the filters with different sizes to aggregate more information, and [20] tries to use the block instead of single average for improving SENet scheme [4]. In [21] three variant squeeze and excitation modules are tested respectively following spatial dimension, channel dimension or both. In addition, [16,19] notice that the fully connected transform using reduction, as in [4], will lead to information loss, and propose to keep the channels number unchanged.

In contrast, the spatial attention mechanism, taken non-local [5] as the typical example, aims at modeling the global dependency by aggregating contextual information as far as possible. So it inevitably increases the total complexity and easily leads to out of memory. The double attention networks [29] reinterpret the non-local from the view of gathering the features from entire space and distributing the features to each location. From the aspect of gathering the features, the works [10,11,12,13] decrease the computing complexity at the cost of restricting the feature gathering within a limited field. In [14] the contextual information gathering is carried out following two stages: the first is for long range attention and the second is for short dependency. In both stages the computing scale is decreased by dividing the feature maps into 4 groups. The dual attention network [7] enhance the non-local by gathering the information not only at all positions but also from all channels, and finally both attentions from spatial and channel dimensions are summed to improve feature representation. It achieves the SOTA performance for scene segmentation at the cost of more computing and memory resources than non-local attention network [5,7].

In addition, some researches consider to gather the contextual information from different ways. In [30] the attention maps are regarded as the summation of a compact set of bases, which can be estimated through the Expectation-Maximization process, and especially the resulting feature representation based on these bases will be low rank. So the contextual information gathering could be viewed as the analysis and synthesis upon the learned bases. [31] proposes a local relation network to exploit the geometry prior information for dealing with varying spatial distribution. After training the local relation network will learn how to gather the contextual information when handling the new data with different appearance, however the need of the source code prevents us from appreciating the model deeply. Recently neural architecture search is applied to non-local block [5] in order to find an optimal configuration for mobile neural network [15]. During the contextual information gathering, the proposed lightNL block [15] down-samples the input feature maps along either channel dimension, spatial dimension or both dimensions, so the computing burden will be reduced. In addition, it uses a shared transformation generate the query, key and value, and the reusing of transformation will further decrease the computation cost. An interesting work comes from [22]. Based on the observation that the attention maps are approximately invariant to changes in query positions, GCNet [22] models the contextual information via one 1×1 convolution by transforming the input feature maps' channels into only one, and in nature this transform averages the feature maps in channel level instead of in each spatial position. The work [35] also reports the above observation, and improves the attention by integrating the embedding positions into the calculation of query-key scores.

## 2.3 SPANet vs. Non-local vs. GCNet vs. lightNL

Here we specifically discuss the difference among the proposed SPANet, Non-local [5], GCNet [22], and lightNL [15]. Non-local network [5] generates the attention maps for each query position, and the attention maps gather information from all spatial positi-ons, and then distribute the gathered information to each spatial position [29]. In contrast, the other three methods don't gather contextual information from all spatial positions. GCNet [22] gathers information in the level of channel, instead of each position. lightNL [15] tries to uniformly down-sample the feature maps following the spatial or/and channel dimension. Different from them, the proposed SPANet selectively gathers contextual information from the salient positions, and the position number could be controlled according to a hyper-parameter. From the view of global dependency, the proposed SPANet tries to exclude the unrelated information which is inevitably presented especially for images with complex scene.

From the viewpoint of transformations, these methods work differently. Non-local network [5] uses three different transforma-tions to generate query, key and value respectively, GCNet [22] only use one transformation to generate a mask with only one channel, and lightNL[15] uses the shared transformation to generate query, key and value. Different from them, the proposed work SPANet use the shared transformation to generate query and key, but uses one different transformation for value.

At last, Non-local network [5] gathers and distributes the information both in spatial dimension, lightNL [15] also works the same way, and GCNet [22] collects information from the channel dimension. Different from them, the proposed SPANet selects the salient positions at the spatial dimension, but distribute the conte-xtual information at the channel dimension, which will be des-cribed in Sec.3.1.

## 3 The Proposed SPANet

In this section we will firstly review the core concepts of the self-attention [3] and illustrate some observations that motivated the SPANet in Sec.3.1, and then in Sec.3.2 introduce the details of the proposed SPANet. Last, some discussion about the proposed method will be presented.

### 3.1 Revisiting the Self-attention

The self-attention scheme could be described as below:

$$A = soft\max(\frac{\theta(X)\phi(X)^T}{\sqrt{d}}), \qquad (1)$$

$$Y = Ag(X), \qquad (2)$$

where $X$ and $Y \in R^{n \times c}$ are the input and output feature maps respectively, $A \in R^{N \times N}$ is the affinity matrix, $n$ is the number of spatial positions of the feature map and $c$ is the channel numbers. The transformations $\theta$, $\phi$ and $g$ generate the query, key and value matrices respectively. By multiplying the query with the value matrices transposition followed by the *softmax*, we will get the affinity matrix $A$. In essence each row of the affinity matrix gives out the relation between the corresponding query position and all positions. So, sometimes it is called as the query-key scores. A straightforward example to reveal the mode of self-attention is shown in Figure 1, Figure 2, and Figure 3 from the views of affinity matrices and attention maps respectively. Combined with Figure 1, Figure 2, and Figure 3, some interesting observations can be concluded:

(1) Evidently in the low levels of neural networks, NL-block [5] opts to generate different attention maps in different query positions, while in high levels produce the similar ones, as both Figure 2 and Figure 1 show.

(2) In the high level, the hot areas may be misplaced, while in the low level, the hot areas often occur in the places with rich texture or edge as the top two rows shown in Figure 1.

(3) In comparison with the Figure 3 with clean background, the attention maps in Figure 1 have many hot areas, especially the top two rows. The difference between the attention maps in Figure 1 and Figure 3 implies that not all information be positive for building the contextual model, and the negative one, e.g. the hot areas in the texts of Figure 1 should be excluded.

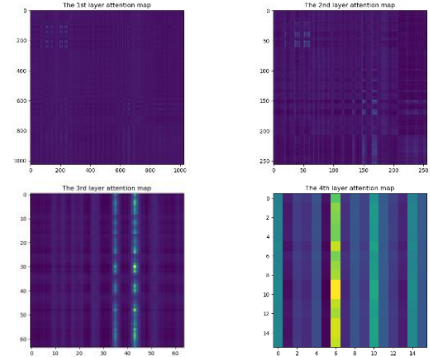

**Figure 2:** The affinity matrix generated by NL-block [5] in ResNet18 [32], from top to bottom and left to right are the maps belonging to stage 1, 2, 3, and 4, with size 1024×1024, 256×256, 64×64, and 16×16 respectively. Each row of the matrix contains the attention map corresponding to the query position same with the row number. The input image is an airplane of CIFAR, same with Figure 1.

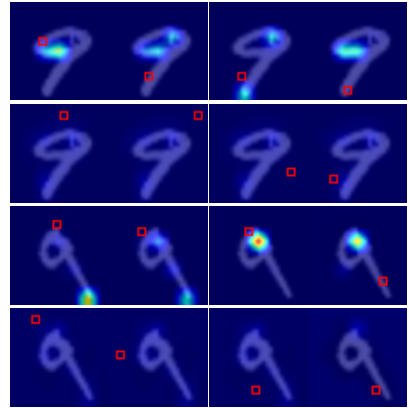

**Figure 3:** The attention maps are generated by NL-block[5], and the red boxes denote the query positions. The top and bottom two rows belong to two different digital 9 respectively, which are from the MNIST datasets [34]. Each attention map has the size of 14×14 and resized to be 56×56 for visualization.

### 3.2 The Proposed SPANet

Figure 4 gives out the illustration of the proposed SPANet, which aims at distilling the positive contextual information while decreasing the computing complexity. The following equations characterize the proposed SPANet attention mechanism,

$$Q = \theta(X), \qquad (3)$$

$$V = g(X), \qquad (4)$$

$$K = S(Q), \qquad (5)$$

$$A = \text{soft}\max(\frac{K^T K}{C}), \quad (6)$$

$$Y = VA, \quad (7)$$

where $Q \in R^{n \times c}$, $K \in R^{k \times c}$, $A \in R^{c \times c}$, and $X, Y \in R^{n \times c}$.

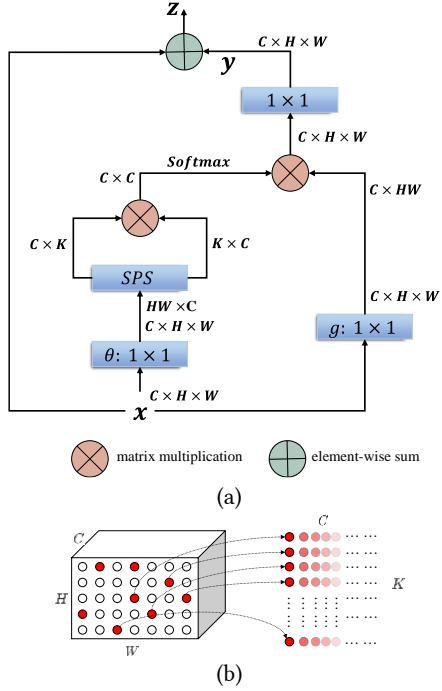

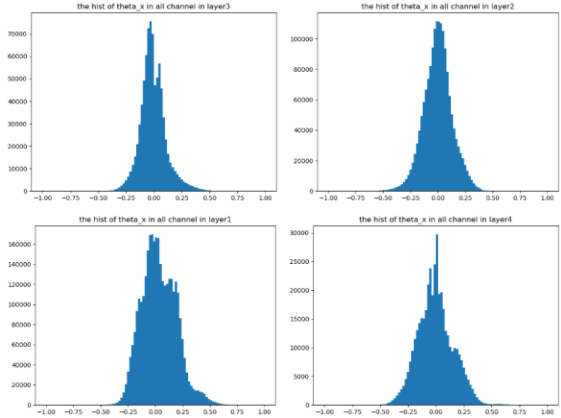

**Figure 4:** (a) The proposed SPANet; (b) The output data structure

**Figure 5:** The histogram of the query matrix data, generated by NL-block [5] in ResNet [32]. From top to bottom and left to right are the maps belonging to stage1, 2, 3, and 4. All data comes from one batch size (=100) of CIFAR [33].

In SPANet, the feature maps firstly pass through two 2d convolution layers to form the query $Q$ and value $V$ matrices (Equ. (3) and Equ. (4)). Then the Matrix $Q$ will be passed to the proposed SPS algorithm, which will be described in Sec. 3.3. The output of SPS module will be the top-k salient positions in query matrix (Equ. (5)). Figure 4(b) illustrates the selection procedure, especially the changes in the dimensions. Equ. (6) uses the selected data from top k positions to compute affinity matrix $A$. In the aggregation process, after the *softmax* of affinity matrix $A$ following the column, the output $Y$ will be calculated by multiplying the $V$ matrices with $A$, and reshaped to be $c \times h \times w$. After a transformation of 1×1 convolution, $Y$ will be finally added to the input $X$.

### 3.3 The Salient Positions Selection Algorithm

In the SPANet the proposed salient positions selection (SPS) algorithm plays an important role. The selection of the salient positions will decrease the computing complexity, and more importantly it tries to distill the positive contextual information during modeling the global dependency. The proposed SPS algorithm is described as below.

| ALGORITHM 1: Salient Positions Selection (SPS) Algorithm |
|---|
| **Input**: matrix $Q$ with size [$c$, $h^*w$], a hyper parameter $k$ |
| **Output**: matrix $K$ with size [$c$, $k$] |
| 1. Computing the square power of $Q^\wedge T$ following the channel dimension. |
| 2. Get $Q_{pow}$ by summing $Q^\wedge T$ following the channel dimension. |
| 3. Select the most biggest $k$ positions from $Q_{pow}$, denoted as *indexk*. |
| 4. Return matrix $K = Q(c, indexk)$ |

### 3.4 Discussion

*3.4.1 The metric used to select the salient positions.* Considering the high dimension of the feature maps, it is not an easy task to select the salient positions, especially we don't exactly know what data in feature maps are important. Why do we choose the sum of squared power of the positions to measure their saliency? The reasons are, at one hand the reported methods good for the visual image are not suitable for the feature maps in neural networks as Sec.2.1 says, and at the other hand the histogram of the feature maps data in all channel follows the approximate Gaussian distribution, as Figure 5 shows. So the squared power of the data will be a favorable metric of measuring their importance under the approximate Gaussian distribution, which apparently differs from the visual images.

*3.4.2 The SPANet computing complexity.* In the whole attention computing process, the main parts come from two multiplication of Equ. (6) and Equ. (7). Considering the size of matrix in the two equations, the total complexity will be $O(c^2 k + nc^2)$. When the SPANet is added in low network levels, the complexity will be greatly decreased, e.g. for the first layer with feature map size 32×32 and 64 channels, the complexity of self-attention will be up $O(2^{26})$, but the proposed SPANet only $O(2^{22})$ when $k = 64$. In the low network level, the computing complexity mainly composes of Equ. (7), and the part from Equ. (6) is negligible. However, as the network level becomes high, the channels will become more and more while the height and width become smaller and smaller, the computing complexity owing to Equ. (6)

and Equ. (7) will be larger than self-attention. So the proposed algorithm will incline to work well in low network well.

*3.4.3 The hyper parameter k.* It is used control the scale of the selected salient positions in query matrix. It is a regret we still don't find a suitable value depending on the theory analysis, and the value used in the source code is set empirically. Based on the coming experimental results we believe that it will be a valuable work to do theory analysis in this question.

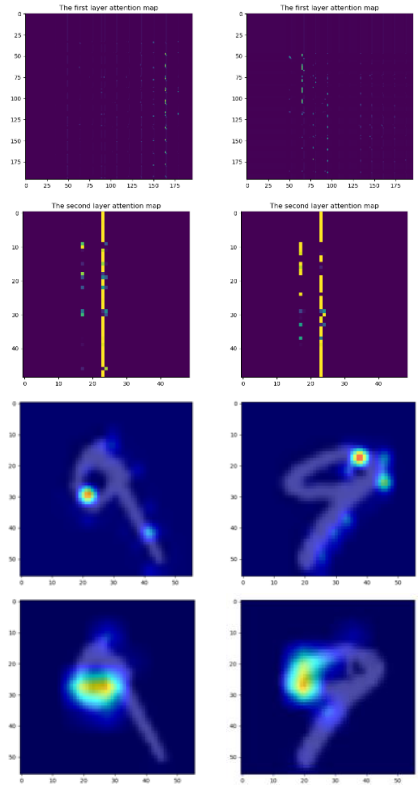

**Figure.6: The examples of attention-maps produced by NL-block [5]. The two digital numbers [34] are same with the two nine in Figure 3. The top two rows are the affinity matrices generated according to Equ. (1) in 1st and 2nd network layer respectively, and each row of them corresponds to an attent-ion map. All the attention maps corresponding to all columns of the top two rows matrices are averaged and resized to 56×56. So the attention maps of the first row are shown as images in the third row, and those of the second row are shown as images in the fourth row.**

## 4 Experimental Results

To evaluate the effectiveness of the SPANet, we carry out various experiments on datasets of MNIST [34], CIFAR [33] and Tiny-ImageNet. In order to compare fairly we don't use any other tricks in train during the experiments, such as positional embedding, extra training data. Experimental results demonstrate that SPANet achieves state-of-the-art performance on above datasets. Because all methods could achieve the top-1 accuracy more than 99% on MNIST [34], we only use the intermedia results of MNIST for analyzing as 4.1 describes. The comparisons use datasets include CIFAR-10, CIFAR-100 and Tiny-ImageNet, which is a subset of ImageNet [37]. All images of Tiny-ImageNet are resized to 64×64 pixels. In the following unless otherwise specified, the backbone networks all use the ResNet [32], and make layers in four stages [38], and the metric of image classification is the top-1 accuracy (in %). Some source codes, such as the implementation of NL-Block, SENet, are borrowed from the repository [38,39,40], and all experiments use the following sets: the batch size is 128, the learning rate is 0.9 with SGD *momentum* = 0.9 and *weight_decay* = 5e-4, and the scheduler uses CosineAnnealingLR.

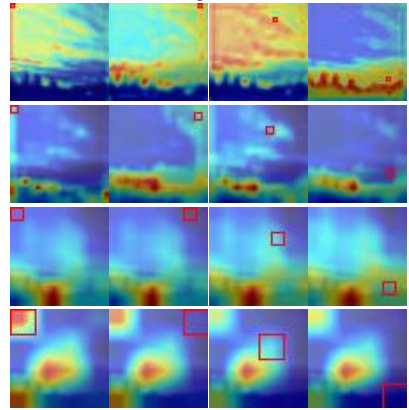

**Figure 7: Some attention maps produced same with Figure 1. The image in CIFAR-100 [33] with label cloud is predicted as sea.**

### 4.1 Some observations in simple and complex images

The MNIST dataset [34] consists of the handwritten digitals from 0 to 9, and there is no a cluttered background to distract the network. Although it is not enough for comparing the various networks performance, we believe it is a good target for analyzing the network inner work mode. Except the observations listed in Sec.3.1, from Figure 6 and Figure 7 we can conclude that:

(1) The affinity matrices and attention maps of the digitals all have the hot spots fall in the objects themselves.

(2) In Figure 6 with a clean background, in the first layer, only the query positions in object could have strong hot spots in attention maps, as shown in Figure 3, while in the second layer, all query positions produce strong hot spots.

(3) In contrast to Figure 6, the images with clutter background in Figure 1, Figure 2 and Figure 7 will generate hot spots in several places instead of only in the recognized object. When the hot spots in the attention maps are misplaced as shown in the bottom row of Figure 7, the prediction will be wrong. These observations may imply that the relation between the query positions and the hot spot positions in attention maps maybe of more importance.

Insert Your Title HereWe notice that although most modification works of self-attention get only approximate performance, [10] achieves better performance than NL-block [5] with lower computing complexity and memory. The authors contribute the reason to the sequentially recurrent operations criss-cross attention. However, based on the above observations we argue that the gain maybe resulted from the criss-cross way of collecting the contextual information beca-use the criss-cross excludes out the clutter background to some extent. The reason why we do such deduction is that the proposed SPANet has the similar performance. Different from CCNet [10] with the fixed selection way of criss-cross, the proposed SPANet selects the salient positions according to the squared power of the input feature maps.

**Table 1: Experimental results of the image classification dataset**

| Network | # of Params | depth | CIFAR-10 | CIFAR-100 | Tiny-ImageNet |
|---|---|---|---|---|---|
| resnet [32] | 11.174M | 18 | 93.07 | 75.87 | - |
|  | 23.521M | 50 | 93.65 | 78.16 | - |
| resnet + SE [4] | 11.268M | 18 | 95.12 | 76.44 | - |
|  | 26.078M | 50 | 95.14 | 78.58 | - |
| resnet +GC [22]* | 11.191M | 18 | 95.38 | 78.38 | 65.98 |
|  | 23.785M | 50 | 95.69 | 79.13 | - |
| resnet +NL [5]* | 11.306M | 18 | 95.36 | 78.39 | 65.64 |
|  | 25.622M | 50 | 95.37 | 79.15 | **69.26** |
| resnet+SPA ($k$=2)* | 11.371M | 18 | **95.79** | 78.68 | 66.44 |
|  | 26.671M | 50 | 95.65 | **79.47** | 69.15 |

Notice: * denotes the module is added to stage 3. – denotes no measure.

**Table 2: Experimental results of different stages, 1 means True and 0 means False in stage column**

| Network | depth | stage | CIFAR-100 |
|---|---|---|---|
| resnet +NL | 18/50 | (1,0,0,0) | 78.50/79.84 |
|  |  | (0,1,0,0) | 78.26/78.90 |
|  |  | (0,0,1,0) | 78.39/79.15 |
|  |  | (0,0,0,1) | 77.92/78.76 |
| resnet +SPA | 18/50 | (1,0,0,0) | 78.51/**80.01** |
|  |  | (0,1,0,0) | 78.59/79.78 |
|  |  | (0,0,1,0) | 78.68/79.47 |
|  |  | (0,0,0,1) | 76.82/78.46 |

## 4.2 Experiments on image classifications

*4.2.1 The Comparisons with other works.* Besides the back-bone network ResNet [39,32], Table 1 gives out the results of the typical channel attention scheme SENet [4] and the dotproduction scheme NL-block [5]. The implementation of SENet [5] is based on [40], and NL-block is based on [38]. Both have a slight better top-1 accuracy than the original ones. In addition, the GCNet, aiming at fusion the both advantages of SENet and NL-block is tested using the source code from [38]. Because of this paper mainly focusing on the effects of distilling positive information in contextual modeling, the NL-block, taking all contextual information into account, will be selected as the major reference.

The results of Table 1 show that, at a cost of a slight para-meters increasing, the proposed SPANet improves the backbone network performance more than 2 percentages, no matter in cases of 18 or 50 depth. The results hold true for CIFAR-10, CIFAR-100 and Tiny-ImageNet. Because the convolutions in Equ. (3) and Equ. (4) is more complex than Equ. (5) and Equ. (6) in SPANet, the computing complexity decreasing resulted from Equ. (5) and Equ. (6) becomes not so much important, but the spared memory in comparison with NL-block cannot be overlooked. In general, the proposed SPANet achieve a better performance than NL-block with a lot of memory saving and a partial reduction of computing complexity.

*4.2.2 which stage to add SPANet?* In above section it can be concluded that the proposed SPANet and NL-block in general acquire the top-2 performance. So in the following experiments we will focus the comparisons between SPANet and NL-blocks, and takes CIFAR-100 as the test targets.

Table 2 shows the classification results of adding SPANet in different network stages. Two facts could be found that:

(1) From the 1$^{st}$ stage to 3$^{rd}$ stage, no matter SPANet and NL-block can improve the backbone network performances more than 2 percentages, and the SPANet outperforms the NL-blocks, e.g. when $k$ = 8 for ResNet18 and $k$ = 4 for ResNet50.

(2) In the 4$^{th}$ stage, the last stage of backbone network with the smallest feature map size, the performances of SPANet and NL-block, although slightly better than the backbone network, are inferior to themselves in the lower network levels.

Above results not only prove that the proposed SPANet can improve the image classification accuracy, but also demonstrate that the suitable application scope of SPANet is in the low and middle network levels. Further, combining with Figures 1 to 3, Figure 6 and Figure 7, we can conclude that the proposed SPANet does distill the positive contextual information, and modeling glo-bal dependency. In contrast, for the global dependency modeling in NL-block [5] it is difficult to avoid all noisy information especially when there is clutter background, and inevitably this has NL-block [5] suffered from a performance loss.

In addition, the results of the 4$^{th}$ stage show both methods suffer a performance degradation, which has been reported in [5] and thought to be the need of precise spatial information. However, we argue the degradation may also come from that the negative info, extracted from the clutter background in the low level, are amplified in the last network level. The reasons we do such deduction are that: in the shallow network levels the atten-tion maps show the various patterns, in contrast in the deep netw-ork levels the attention maps tend to be similar. How to selecti-vely gathering the positive information in the low level and stopp-ing the error drift between network layers need further research.

*4.2.3 The effect of the hyper-parameter $k$.* In essence the value of $k$ is to control the scale of the salient positions selected in the SPS algorithm. Evidently, a larger $k$ will allow more positions to attend the affinity matrix at the cost of more memory. Although a small $k$ will result in less burden, we would worry about whether the positions are enough for producing good attention maps. At first, we set the reduction ratio (=16) to adaptively select the positions according to the feature map size for each network stage. However, the experiment results surprise us and make us set the $k$ empirically. From Table 2 we can conclude that:

(1) The proposed SPANet can achieve better or equivalent performances than NL-block. For example, when $k$ = 2 SPANet outperforms NL-block in most stages. Under this condition, the selected positions are surprisingly less than one percentage of NL-block, especially when in the 1st and 2nd stages.

(2) The large $k$ cannot ensure a better performance than small $k$.

(3) Which value of $k$ will be good for different stage or backbone network are far from conclusive.

Above observations, especially the performance of SPANet with $k$ = 2, imply that the problem of salient positions selection is worth more researches, especially combining with the knowledge distilling [42].

In addition, we try to add the SPANet simultaneously into two different stages in backbone network, and present the results in Table 3. However, in comparison with only one stage, the extra SPANet leads to only little gain. At some cases it even suffers from a degradation. In fact, more NL-blocks also fall into this case as shown in Table 3. These results demonstrate the import-ance of the research in the working mode between network layers from another aspect.

**Table 3: Experimental results of different hyper-parameter $k$, tested on ResNet18, – denotes no measure**

| $k$ \ stage | (1,0,0,0) | (0,1,0,0) | (0,0,1,0) | (0,0,0,1) |
|---|---|---|---|---|
| $k$ = 2 | 78.51 | 78.59 | **78.68** | 76.82 |
| $k$ = 4 | 78.47 | 78.41 | 78.18 | 76.59 |
| $k$ = 8 | **78.52** | **78.86** | 78.44 | - |
| $k$ = 16 | 78.25 | 77.97 | 78.50 | - |
| $k$ = 32 | 78.31 | 77.02 | - | - |

## 5 Conclusion and Future Works

In this paper we present a salient positions-based attention mech-anism, SPANet, which decreases the computing complexity on the one hand, and on the other hand tries to distill the positive inform-ation when gathering the contextual information. Owing to both benefits, the proposed SPANet shows better image classification accuracies in many experiments with only a fraction complexity of self-attention. In addition, some interesting observations on the affinity matrices and attention maps generated on different netwo-rk layer and datasets are presented and discussed. For future work, we will explore how to learn the suitable salient positions by the network itself instead of controlling it by a hyper-parameter. In addition, we also are interested in the inner mode of the affinity matrices in different network layers and the theory of distilling the positive contextual information. At last, more extensive experim-ents on various datasets will be carried out.

# Insert Your Title Here